\documentclass{svmult}

\usepackage{dream}

\usepackage{helvet}

\usepackage{courier}

\usepackage{graphicx}
\usepackage{makeidx}
\usepackage{multicol}
\usepackage{footmisc}

\usepackage{xspace}

\begin{document}
\title*{Adventures in Mathematical Reasoning}

\author{Toby Walsh}
\institute{University of New South Wales, Sydney and Data61}

\maketitle

\begin{abstract}
{\em ``Mathematics is not a careful march down a well-cleared highway, but a journey into a strange wilderness, where the explorers often get lost. Rigour should be a signal to the historian that the maps have been made, and the real explorers have gone elsewhere.''} W.S. Anglin, the Mathematical Intelligencer, 4 (4), 1982.
\end{abstract}

\newcommand{\set}{\mathcal}
\newcommand{\myset}[1]{\ensuremath{\mathcal #1}}

\renewcommand{\theenumii}{\alph{enumii}}
\renewcommand{\theenumiii}{\roman{enumiii}}
\newcommand{\figref}[1]{Figure \ref{#1}}
\newcommand{\tref}[1]{Table \ref{#1}}
\newcommand{\myldots}{\ldots}

\newtheorem{myproblem}{Problem}
\newtheorem{mydefinition}{Definition}
\newtheorem{mytheorem}{Theorem}
\newtheorem{myobservation}{Observation}
\newtheorem{mylemma}{Lemma}
\newtheorem{mytheorem1}{Theorem}
\newcommand{\myproof}{\noindent {\bf Proof:\ \ }}
\newcommand{\myqed}{\mbox{$\Box$}}
\newcommand{\myend}{\mbox{$\clubsuit$}}

\newcommand{\mymod}{\mbox{\rm mod}}
\newcommand{\mymin}{\mbox{\rm min}}
\newcommand{\mymax}{\mbox{\rm max}}
\newcommand{\range}{\mbox{\sc Range}}
\newcommand{\roots}{\mbox{\sc Roots}}
\newcommand{\myiff}{\mbox{\rm iff}}
\newcommand{\alldifferent}{\mbox{\sc AllDifferent}}
\newcommand{\permutation}{\mbox{\sc Permutation}}
\newcommand{\disjoint}{\mbox{\sc Disjoint}}
\newcommand{\cardpath}{\mbox{\sc CardPath}}
\newcommand{\CARDPATH}{\mbox{\sc CardPath}}
\newcommand{\common}{\mbox{\sc Common}}
\newcommand{\uses}{\mbox{\sc Uses}}
\newcommand{\lex}{\mbox{\sc Lex}}
\newcommand{\usedby}{\mbox{\sc UsedBy}}
\newcommand{\nvalue}{\mbox{\sc NValue}}
\newcommand{\slide}{\mbox{\sc CardPath}}
\newcommand{\sliden}{\mbox{\sc AllPath}}
\newcommand{\SLIDE}{\mbox{\sc CardPath}}
\newcommand{\circularslide}{\mbox{\sc CardPath}_{\rm O}}
\newcommand{\among}{\mbox{\sc Among}}
\newcommand{\mysum}{\mbox{\sc MySum}}
\newcommand{\amongseq}{\mbox{\sc AmongSeq}}
\newcommand{\atmost}{\mbox{\sc AtMost}}
\newcommand{\atleast}{\mbox{\sc AtLeast}}
\newcommand{\element}{\mbox{\sc Element}}
\newcommand{\gcc}{\mbox{\sc Gcc}}
\newcommand{\gsc}{\mbox{\sc Gsc}}
\newcommand{\contiguity}{\mbox{\sc Contiguity}}
\newcommand{\PRECEDENCE}{\mbox{\sc Precedence}}
\newcommand{\assignnvalues}{\mbox{\sc Assign\&NValues}}
\newcommand{\linksettobooleans}{\mbox{\sc LinkSet2Booleans}}
\newcommand{\domain}{\mbox{\sc Domain}}
\newcommand{\symalldiff}{\mbox{\sc SymAllDiff}}
\newcommand{\alldiff}{\mbox{\sc AllDiff}}

\newcommand{\slidingsum}{\mbox{\sc SlidingSum}}
\newcommand{\MaxIndex}{\mbox{\sc MaxIndex}}
\newcommand{\REGULAR}{\mbox{\sc Regular}}
\newcommand{\regular}{\mbox{\sc Regular}}
\newcommand{\precedence}{\mbox{\sc Precedence}}
\newcommand{\STRETCH}{\mbox{\sc Stretch}}
\newcommand{\SLIDEOR}{\mbox{\sc SlideOr}}
\newcommand{\NAE}{\mbox{\sc NotAllEqual}}
\newcommand{\mytheta}{\mbox{$\theta_1$}}
\newcommand{\mysigma}{\mbox{$\sigma_2$}}
\newcommand{\mysigmatwo}{\mbox{$\sigma_1$}}

\newcommand{\todo}[1]{{\tt (... #1 ...)}}
\newcommand{\myOmit}[1]{}

\newcommand{\dpsb}{DPSB}

\section{Introduction}

In 1986\footnote{How can it be that long ago?}, I moved
to Edinburgh to start a Masters conversion course into
Artificial Intelligence after having studied mathematics
at the University of Cambridge. I had dreamed about working in AI
for many years. So it was my good fortune to fall into the
gravitational attraction of Alan Bundy and become 
a member of the DReaM group. Shortly after, I began a PhD
under Alan's careful supervision\footnote{I was lucky
also to have one of Alan's postdocs, Fausto Giunchiglia as a second supervisor. We would work together closely for
the next decade.} \cite{phd,gwjai,theories}. 

I would now
start to dream about getting
computers to do mathematics. It was perhaps not surprising
that this was the orbit into which I fell. I had
always liked mathematics and now I could combine two
of my passions: Artificial Intelligence and mathematics. 

I would stay in Edinburgh for most of the next dozen or so
years, apart from some enjoyable excursions to
work at INRIA in Nancy and with Fausto Giunchiglia at IRST in Trento. 
There was a lot to
like about living and working in Aulde Reekie. 
However, fresh challenges started to emerge and 
I began to build up to
an escape velocity. I took 
a research position in Glasgow but stayed
living in Edinburgh, visitng the DReaM group
frequently. Then I moved to York, and finally
a sling shot sent me past Cork to Sydney, Australia
where I remain today. 

In due course, I would leave behind the work that I had done in
Edinburgh and explore other parts of Artificial Intelligence. However, 
this volume offers me the chance to  consider how those ripples
might have had a small influence on what followed.
More importantly, it lets me thank Alan for his mentoring. 
This summary is necessarily high level and will
avoid going into many of the technical details. I have
a lot of ground to cover so it is impossible in
this limited space to go deeper. 

\section{Rippling}

Alan was (and is) a neat AI researcher. 
One strategy he promoted was to take some
scruffy research and make it neat. He even gave
it a name: undertaking a ``rational reconstruction'' of
some past research.  In the late
1980s \cite{rational}, the DReaM group embarked on 
a project to rationally
reconstruct the rather scruffy
inductive theorem proving techniques to be
found in Boyer and Moore's NQTHM theorem prover
\cite{boyer1}. Dieter Hutter in Karlsruhe was 
going about a similar task developing the INKA 
prover \cite{hutter} and soon became closely involved
with the efforts in Scotland. 

Central to the inductive theorem proving heuristics in NQTHM were
the ideas of recursion analysis
(picking an induction rule and 
variable), and then rewriting the step case
to match the induction hypothesis in a process
that became rationally reconstructed 
as an annotated form of rewriting 
called ``rippling'' \cite{pub459,pub567}.
Picking an appropriate induction rule and variable
depends on how you can simplify the resulting 
step case so the two ideas are closely 
connected. 

In the early 1990s, rippling ran into some annoying
problems. In particular, the annotations used by rippling to guide
rewriting could become ill-formed, and there was
as yet no principled way to annotate terms in the 
first place. Alan's newest postdoc, David Basin and I 
set about fixing these problems. I shall tell the story
backwards as it makes for a more rational reconstruction
of the work. 

\subsection{A calculus for rippling}

Rippling is a form of rewriting guided by special kinds of
(meta-level) annotation. Consider, for example, a proof 
that appending a list onto the empty list 
leaves the list unchanged. 
In the step-case, the induction hypothesis is, 
\begin{eqnarray*}
(append ~ x ~ nil) & = & x
\end{eqnarray*}
From this, we need to derive the induction conclusion,
\begin{eqnarray*}
(append ~ (cons ~ e ~ x) ~ nil) & = & (cons ~ e ~ x)
\end{eqnarray*}
We can annotate the induction conclusion to highlight
the differences between it and the induction hypothesis,
\begin{eqnarray*}
(append ~ \wf{(cons ~ e ~\wh{x})} ~ nil) & = & \wf{(cons ~ e ~ \wh{x})}
\end{eqnarray*}
The square boxes are the ``wavefronts''. The underlined parts are the 
``waveholes''. If we eliminate everything in the wavefronts but not in 
the waveholes, we get the ``skeleton''
in which the induction conclusion matches exactly
the induction hypothesis. 

To simplify the induction hypothesis, we will use a rewrite
rule derived from the recursive definition of append,
\begin{eqnarray*}
(append ~ {(cons ~ a ~{b})} ~ c)& =_{\rm def} & 
{(cons ~ a ~ {(append ~ b ~ c)})}
\end{eqnarray*}
We can turn this into a rewrite rule annotated with wavefronts and
waveholes,
\begin{eqnarray*}
(append ~ \wf{(cons ~ a ~\wh{b})} ~ c)& \Rightarrow & 
\wf{(cons ~ a ~ \wh{(append ~ b ~ c)})}
\end{eqnarray*}
This is called a ``wave rule''. It preserves the skeleton,
$(append a c)$ but moves
the 
wavefront, $(cons a \ldots)$ up and hopefully out of the way.  
 Applying this rule to the left-hand side of the induction conclusion gives,
\begin{eqnarray*}
\wf{(cons ~ e ~\wh{(append ~ x ~ nil)})}& = & \wf{(cons ~ e ~ \wh{x})}
\end{eqnarray*}
We can now simplify the left-hand side using the induction hypothesis as
the rewrite rule, 
\begin{eqnarray*}
(append ~ x ~ nil) & \Rightarrow & x
\end{eqnarray*}
This rewriting step
is called fertilization, and leaves the following equation, 
\begin{eqnarray*}
\wf{(cons ~ e ~\wh{x})}& = & \wf{(cons ~ e ~ \wh{x})}
\end{eqnarray*}
The left-hand side of the rewritten induction conclusion 
now matches the right-hand side. The step case therefore holds 
by the definition of equality. 

Rewriting annotated terms in this way take us beyond the normal rewriting of
terms. David and I therefore formalized a calculus for
describing such annotated rewriting \cite{calculus-of-rippling,jar-rippling}. 
We showed that this calculus had the following 
four desirable properties.
\begin{description}
\item[Well-formedness:]  rewriting properly annotated terms
with wave rules leaves them properly annotated.
\item[Skeleton preservation:] 
rewriting properly annotated terms
with wave rules preserves the skeleton. 
\item[Correctness:] 
we can perform the corresponding
derivation in the underlying un-annotated theory.
Annotation is thus merely guiding search.
\item[Termination:] 
given appropriate measures on annotated terms, we can guarantee 
rippling terminates. There are, for example, no 
loops. \end{description}
We showed that different termination orderings can 
profitably used within and outwith induction
\cite{termination-rippling}. Such new orderings let us combine
the highly goal directed features of rippling with the 
flexibility and uniformity
of more conventional term rewriting. 
For instance, we proposed two new orderings which allow unblocking,
definition unfolding, and mutual recursion to
be added to rippling in a principled (and terminating) fashion. 

\subsection{Difference matching and unification}

But where do the annotations come from in the first place? 
David and I generalized both (1-sided) matching and
(2-sided) unification to annotate terms appropriately. 
Difference matching extended first-order matching to make one term,
the
pattern 
match another, the target by instantiating variables in the target, 
as well as by hiding structure with wavefronts also in the target \cite{pub556}.
Difference unification extended unification to make
two terms syntactically equal 
by variable instantiation and by hiding
structure with wavefronts in both terms
\cite{BasinWalshDu}. Difference unification is needed,
for instance, to annotate rewrite rules as wave rules. 

A single difference match can be found in time linear
in the size of the target. If the pattern contains a variable,
then set this to the target and put everything else in the wavefront.
If not, the pattern is ground and we can simply descend through the term
structure hiding any differences between the pattern and the target in
wavefronts. However, there can be exponentially many
difference matches in the size of the pattern in general so returning all of
them can take exponential time. In practice, though, 
there are usually only a few successful difference matches
and these can be found quickly.

Difference unification is more problematic computationally. Even if we limit
wavefronts to one term, deciding if two terms difference unify 
together is NP-hard (Theorem 8 in \cite{BasinWalshDu}). 
Thus, supposing P $\neq$ NP, we cannot
in general find even a single difference unifier in polynomial time. 
Looking again at this result more than 25 years later, I would
not leave the analysis there but would look closer at the source of
complexity. Difference unification hasn't proved 
too intractable in practice and we can likely argue why not. 
The reduction showing NP-hardness
reduced a propositional satisfiability
problem in $m$ clauses to difference unifying
two $m$-ary functions. The functions being 
difference unified in this proof therefore 
can have very great arity. I conjecture that
difference unification
is polynomial, more precisely
fixed-parameter tractable, when applied to terms of bounded arity. 

~ \\

\framebox[0.9\textwidth]{%
 \begin{minipage}{0.85\textwidth}
{\bf An unexpected tale:}

\vspace{0.5em}

To find the difference unification with the least amount of 
annotation, we proposed
a new generic AI search called left-first search (LFS)
\cite{BasinWalshDu}. Left branches of 
our search tree introduced annotations, whilst right branches
matched terms. Left-first search explored leaf nodes of this search tree
in order of the number of left branches taken. I presented
the search method at IJCAI 1993. 

\vspace{0.5em}

Two years later, I was listening to an IJCAI 1995 conference talk on
a new search method called limited discrepancy search (LDS) \cite{lds}
when one of the authors put up a slide showing the 
order of the leaf nodes explored by LDS. This 
appeared identical to that of LFS, a slide I remember preparing
two years before. At the end of the talk, I therefore asked
about the difference between the two search methods. A colleague
called it the ``question from hell'' but my intention 
was just to understand how they differed. 

\vspace{0.5em}

Unbeknown to me, LDS was being patented, and 
it set off a chain of unfortunate events. Lawyers
had to rewrite the patent application at some significant 
cost. I was asked to be
an expert witness in a patent dispute over LDS. And
I was considered by the authors of LDS to be a ``trouble maker''. 

\vspace{0.5em}
 
Eventually it
would blow over as there is a simple but
crucial difference. Our search trees were small and
so LFS expanded them in memory. LDS was intended for
much larger search trees and so, whilst it expanded
leaf nodes in the same order as LFS, did so in a lazy
space efficient fashion by returning repeatedly to the root node
much like iterative deepening search. This adds just a constant
factor to the time asymptotically so is worth paying when
space is an issue.

\vspace{0.5em}

  \end{minipage}
}

~ \\

Whilst difference unification was invented to 
deal with inductive proof, it captures a deeper and more 
general idea used in mathematics. 
In \cite{robunify}, J.A. Robinson presented a simple account of 
unification in terms of difference reduction. He
observed,
\begin{quote}
{\em ``Unifiers remove differences ... We repeatedly reduce the difference between the two given
expressions by applying to them an arbitrary reduction of the difference and accumulate the
product of these reductions. This process eventually halts when the difference is no longer
negotiable [reducible via an assignment], at which point the outcome depends on whether the
difference is empty or nonempty.''}
\end{quote}
Difference unification can be seen as a direct extension of Robinson's notion of difference reduction:
we reduce differences not just by variable assignment, but also by
term structure annotation. However, what makes
this extended notion of unification attractive, is that this annotation is precisely what is
required for rippling to remove this difference. And, as we see
shortly, rippling has found an useful role to play in a number of other
proof areas. 

\section{Proof planning}

An important idea explored within the DReaM group
is the separation of logic and control. Proof planning,
developed originally for inductive proof \cite{proofplan}, brought together ideas of
meta-level control explored in the earlier PRESS project \cite{press}
with AI planning operators specified by pre- and post-conditions, 
Theorem proving heuristics are described by general
purpose proof planning 
methods such as rippling and fertilization that are glued together using
simple AI search techniques like depth-first or best-first search. 
Since proof planning was proving useful for inductive proof,
I became keen to try to apply it elsewhere.

\subsection{Summing series}

To explore the use of proof planning in general,
and rippling in particular outside of inductive proof, 
I chanced on the domain of 
summing series \cite{sumseries}.
Inductive proofs can be used to verify identities
about finite sums. But where do these identities 
come from in the first place? 

I developed a set of proof planning methods
to solve such problems. 
To my surprise, rippling proved to be key to many
of these methods. I will illustrate this with
the {\sc Conjugate} method. 
This method transforms a finite sum of terms into the finite sum of some
conjugate. The conjugate can be one of several second
order operations, e.g. the differential or integral of the original term, or the mapping
of a trigonometric series onto the real or imaginary part of a complex
series. 

Consider, for example, find a closed form expression for,
$$ \sum_{i=0}^n  (i+1)x^i $$
The {\sc Conjugate} method transformed this into a simpler looking 
sum,
$$ \sum_{i=0}^n  \frac{{\rm d} x^{i+1}}{{\rm d}x} $$
This now looks close to a 
known result, the closed form sum of a geometric series,
\begin{eqnarray*}
 \sum_{i=0}^n  x^i & = & \frac{x^{n+1}-1}{x-1}
\end{eqnarray*}
Difference matching our simpler looking sum
against the left-hand side of this know result
gives some wavefronts we need to 
remove out of the way by rippling with wave rules,
$$ \sum_{i=0}^n  \wf{\frac{{\rm d} \wh{x^{\wf{\wh{i}+1}}}}{{\rm d}x}} $$
Since the derivative of a sum is the sum of the derivatives, rippling gives,
$$   \wf{\frac{{\rm d} \wh{\sum_{i=0}^n x^{\wf{\wh{i}+1}}}}{{\rm d}x}} $$
Rippling with a wave rule derived from the definition
of exponentiation then expands the exponent,
$$   \wf{\frac{{\rm d} \wh{\sum_{i=0}^n \wf{x . \wh{x^{i}}}}}{{\rm d}x}} $$
One final rewriting step uses rippling to 
move the constant term outside the sum,
$$   \wf{\frac{{\rm d} ~ x . \wh{\sum_{i=0}^n { {x^{i}}}}}{{\rm d}x}} $$
The {\sc Fertilize} method substitutes the closed form sum for the
geometric series,
$$   \wf{\frac{{\rm d} ~ x . \wh{\frac{x^{n+1}-1}{x-1}}}{{\rm d}x}} $$
Finally a {\sc Differentiate}  method then symbolically computes
a closed form answer by algebraically differentiating the quotient.
The derivation is now complete. 

We subsequently looked at some other
mathematical domains such as theorems about limits 
\cite{colrippling}. 
Rippling and proof planning again proved up to
the challenge. 

\subsection{A divergence critic}

Proof planning methods come with high expectations of success.
Their failure can therefore be a useful tool in patching proofs. 
I explored how the failure of rippling can be used to 
suggest missing lemma needed to complete a proof
by means of  a ``divergence critic''
\cite{divergence-critic,divergence-critic2}. Other
members of the DReaM group
have explored similar ideas in closely related
settings (e.g. \cite{critic,irelandfailure}). 

My divergence critic identified when a proof attempt is diverging
by means of difference matching. The critic then proposed lemmas and
generalizations of these lemmas
to try to allow the proof to go through without divergence.
For example, when the prover failed to show inductively that 
$(rev ~ (rev ~ x)) = x$, the
critic proposed the key lemma, a missing wave rule
needed to complete the proof,
\begin{eqnarray*}
(rev ~ \wf{(append ~ \wh{X} ~ (cons ~ Y ~ nil))}) & = & \wf{(cons ~ Y
                                                        ~ \wh{(rev ~
                                                        X)}}
\end{eqnarray*}

In my view, such failure is something we still exploit too little
in automating mathematical reasoning. As a mathematician,
I spend most of my time failing to prove conjectures. 
But those failures eventually often lead me to find either a proof when
the conjecture is true, or a counter-example when it is false.

\section{Mathematical discovery}

Mathematics is more than just proving theorems. It's 
also defining theories. Inventing definitions. Conjecturing results. 
Finding counter-examples. Developing proof methods. 
And more. One of the pleasures of the DReaM group was
to witness and contribute to automating 
some of these other mathematical activities.

In 1996, Alan started to supervise a young and
ambitious PhD student, Simon Colton. 
I was lucky enough to help out. Simon 
wanted to build a system to invent new theories.
Doug Lenat had shown the possibility to 
do this with AM and the followup Eurisko
\cite{ameurisko,ameurisko2}. 
Simon set out to rationally reconstruct Lenat's
work in his HR program \cite{colton1,cbwaaai2000,cbwicml2000,cbwijh2000}.
This was named appropriately
after the famous double act, Hardy and Ramanujan. 
Actually, HR wasn't much of a rational 
reconstruction of AM other than to work on the same
problem as AM, and to use a two letter name
like AM. 

In a wonderful example of why PhD students shouldn't
listen to their supervisors, I suggested to Simon to keep
well away from number theory. I reasoned that there had
been thousands of years of attention to number theory. 
New and automated mathematical discoveries were more 
likely therefore to be found in some newer and little studied
theory like that of Moufang loops. Fortunately, Simon
ignored this advice and HR made a number of discoveries
in number theory. 
On the other hand, in a wonderful example of why PhD students should
listen to their supervisors, Simon wasn't keen to 
submit an update on his work to AAAI 2000. I persuaded
him to do so, and the paper won the Best Paper award. 

\section{The meta-level}

One of the other rewards of working in the DReaM group was
Alan's attention to the meta-level. This wasn't just the
meta-mathematical level, but the meta-level of doing research.
Alan thought long and hard about how we do research, and how you
could do it better. I still recommend the Researcher's Bible that
Alan co-wrote to my PhD students whether they're starting out, or writing 
up their thesis \cite{bible}. 
And when I left Edinburgh, I ``borrowed'' many
of his techniques for doing research on my own: writing 
half formed ideas down in internal notes, 
trying to think of questions to ask at every 
seminar, giving informal talks on any interesting papers
I'd seen at summer conferences, etc. 

\section{Conclusions}

Out of interest, I downloaded one of Alan's latest paper \cite{bundy2019}. 
To my surprise and pleasure, it repeats and expands on many of 
the ideas I've discussed here. It applies rippling to a new domain,
invariant preservation proofs. The meta-level guidance 
rippling provides is used to build proof patches to recover failed attempt and eventually
finish the proofs. And the paper ends with an appendix containing a formal
definition of rippling, along the lines of the calculus we
presented 25 years ago. It feels just like yesterday.
Thank you for everything, Alan.

\section*{Acknowledgments}

Funded by the European Research Council under the Horizon 2020 Programme via
the Advanced Research grant AMPLify 670077.

\bibliographystyle{splncs}
\bibliography{/Users/tw/Documents/biblio/a-z,/Users/tw/Documents/biblio/a-z2,/Users/tw/Documents/biblio/pub,/Users/tw/Documents/biblio/pub2}

\end{document}